\pdfoutput=1
\documentclass[11pt]{article}

\usepackage[final]{acl}
\usepackage{listings}
\usepackage[most]{tcolorbox}
\usepackage{times}
\usepackage{latexsym}
\usepackage{amsmath}
\usepackage{stfloats}
\usepackage[T1]{fontenc}

\usepackage[utf8]{inputenc}

\usepackage{microtype}

\usepackage{inconsolata}

\usepackage{graphicx}

\usepackage{listings}
\usepackage{tcolorbox}
\usepackage{float}

\title{Evolution in Simulation: AI-Agent School with Dual Memory for High-Fidelity Educational Dynamics}

\usepackage{fontawesome5}

\author{
 \textbf{Sheng Jin\textsuperscript{1,*}},
 \textbf{Haoming Wang\textsuperscript{2, *, \faIcon{envelope}}},
 \textbf{Zhiqi Gao\textsuperscript{3}},\\
 \textbf{Yongbo Yang\textsuperscript{4}},
 \textbf{Bao Chunjia\textsuperscript{5}},
 \textbf{Chengliang Wang\textsuperscript{2, \faIcon{envelope}}} \\
 \textsuperscript{1}Guanghua Law School, Zhejiang University, Hangzhou, China \\
 \textsuperscript{2}Faculty of Education, East China Normal University, Shanghai, China \\
 \textsuperscript{3}School of Data Science, The Chinese University of Hong Kong, Shenzhen, China \\
 \textsuperscript{4}Department of Electrical and Computer Engineering, University of California San Diego, USA \\
 \textsuperscript{5}Institute of Systems Science, National University of Singapore, Singapore \\
 \small{
   \textbf{Correspondence:} \href{mailto:wfrank0222@gmail.com}{wfrank0222@gmail.com}, \href{mailto:wcledutech@gmail.com}{wcledutech@gmail.com}
 }
}

\begin{document}
\maketitle

{
\renewcommand{\thefootnote}{}
\footnote{* These authors contributed equally to this work.}
\footnote{\faIcon{envelope} Corresponding authors.}
}

\begin{abstract}
    Large language models (LLMs) based Agents are increasingly pivotal in simulating and understanding complex human systems and interactions. We propose the AI-Agent School (AAS) system, built around a self-evolving mechanism that leverages agents for simulating complex educational dynamics. Addressing the fragmented issues in teaching process modeling and the limitations of agents performance in simulating diverse educational participants, AAS constructs the Zero-Exp strategy, employs a continuous "experience-reflection-optimization" cycle, grounded in a dual memory base comprising experience and knowledge bases and incorporating short-term and long-term memory components. Through this mechanism, agents autonomously evolve via situated interactions within diverse simulated school scenarios. This evolution enables agents to more accurately model the nuanced, multi-faceted teacher-student engagements and underlying learning processes found in physical schools. Experiment confirms that AAS can effectively simulate intricate educational dynamics and is effective in fostering advanced agent cognitive abilities, providing a foundational stepping stone from the "Era of Experience" to the "Era of Simulation" by generating high-fidelity behavioral and interaction data.
\end{abstract}

\section{Introduction}
Large Language Models (LLMs) have demonstrated exceptional performance across a variety of tasks \cite{jing_2024_what, zhu_2024_large}, such as code instruction, information retrieval, and complex problem-solving. As the capabilities of LLM-based agents continue to advance, researchers have begun exploring the simulation of human behaviors to construct complex systems based on real-world scenarios \cite{li_2024_agent, park_2023_generative}. Such simulations are instrumental in understanding human decision-making processes, developing novel human-computer interaction systems, and driving societal model transformations \cite{wang_2024_education}.  

Among these domains, education is particularly eager to leverage agents to achieve adaptive learning and optimize teaching models \cite{jing_2023_research,wang_2024_factors}. Existing research has already developed educational agents for tasks such as mathematical formula conversion \cite{swan_2023_math} and classroom interaction simulation \cite{jinxin_2023_cgmi,jing_2024_what}. However, some scholars have pointed out the limitations of current agents in the education field: first, there is a lack of systematic modeling of the teaching process, and second, LLM agents struggle to accurately simulate the behaviors and interactions of diverse participants in educational settings. In view of this, our goal is to enhance the realism and research value of agent simulations within educational settings by facilitating complex, multi-participant interactions.

We propose the AI-Agent School (AAS), a multi-agent system capable of simulating multidimensional dynamic educational scenarios. Central to AAS is our proposed Zero-Exp strategy, which establishes a dual memory base for storing experience and knowledge. This strategy effectively divides both the experience and knowledge repositories into short-term and long-term parts. Within the AAS environment, multi-role agents iteratively update these memory bases through preset behaviors and interaction data, achieving autonomous evolution via the core "experience-reflection-optimization" mechanism. Experimental results demonstrate that AAS successfully simulates multidimensional dynamic learning scenarios, the autonomous evolution of multi-agents within AAS enables the high-fidelity simulation of the complex performance and interactions of diverse roles in realistic educational scenarios. Our framework provides a verifiable technical model and theoretical pathway for the development of educational digital twins and the production of valuable educational interaction data.

The contributions of our work are as follows:

First, we proposed AAS, a Multi-Agent Educational Scenario Simulation System. It is capable of capturing teacher-student relationships, peer interactions, and environmental influences, enabling the simulation of real teaching processes. Compared to existing methods, AAS demonstrates advantages in handling multiple roles, multi-variable dynamics, and temporal evolution.

Second, we designed the Zero-Exp mechanism to address the challenges of data scarcity and role behavior consistency in educational simulations. This mechanism guides agents to evolve from a zero-experience state to expert-level behavior using a small set of initialization parameters. Experiments show that Zero-Exp enables agents to generate interaction patterns consistent with real educational scenarios.

Third, this research pioneers a new paradigm of "Computational Education Science", deeply integrating traditional educational research with AI technologies. It lays the theoretical and technical foundation for next-generation educational systems, teacher training platforms, and educational policy simulation tools, propelling the education field from the "Era of Experience" to the "Era of Simulation".

\section{Related Work}

The concept of educational intelligent agents originates from \citet{skinner_1958_teaching} principle of programmed instruction and the programmed teaching machines he designed. The intelligent tutoring systems developed in the 1970s were early iterations of this idea. According to \citet{hayesroth_1995_an}, a pioneer in the field, an educational agent can be defined as a virtual tutoring role within a learning system that is responsible for dynamically sensing the learning environment, analyzing and inferring learner information, and actively or passively performing assistive actions based on needs.

Although early educational intelligent agents did not gain widespread attention due to technological limitations, by the early 21st century, scholars began to develop educational intelligent agents with practical value \cite{kim_2015_researchbased}. Empirical studies have shown that educational agents can promote deep learning  \cite{baylor_1999_intelligent} and enhance motivation \cite{atkinson_2002_optimizing, moreno_2001_the}. However, at that time, agent systems struggled to achieve natural, human-like interaction \cite{cassell_2001_embodied}, with a primary focus on promoting cognitive processes.

With the development of AI technologies, research on educational intelligent agents has entered a new phase. AI agents have given educational intelligent agents "living souls," enabling them to participate more vividly in educational practices \cite{veletsianos_2005_the, moise_2005_the}. Researchers have also shown great interest in defining the teaching roles of agents, exploring various roles such as tutors, assistants, learning partners, collaborators, competitors, and even troublemakers \cite{baylor_2005_simulating, madni_2008_intelligent, goodman_2003_towards}. Through large-scale data training, AI agents have become more embodied and capable of performing interactive roles such as collaboration, encouragement, and guidance in complex teaching activities \cite{pedersen_2022_ai, dai_2024_effects}. At the same time, AI agents have developed a certain level of emotional empathy, even alleviating the marginalization experiences of some learners in traditional classrooms, providing a more engaging learning experience \cite{veletsianos_2005_the, schroeder_2013_how, kim_2013_gendered}.

In recent years, the rapid development of LLM has further advanced research on educational AI agents \cite{chen_2024_empowering}. Empirical cases have already demonstrated the enormous potential of LLM-supported educational AI agents. For example, Lan and Chen constructed a teaching AI agent based on LLM and applied it to teaching sequence words (or ordered words), achieving promising results \cite{lan_2024_teachers}. Other scholars have introduced LLM-supported AI agents in areas such as programming education and foundational AI knowledge teaching \cite{jin_2024_teach,zhang_2024_simulating}, creating new learning logic and models in the classroom. However, the interaction of a single AI agent is limited, making it difficult to fully realize the potential of LLM. The AI agent town developed by the Stanford team has validated this point \cite{park_2023_generative}. In light of this, this study will build upon the logic of the AI agent town and Agent Hospital \cite{li_2024_agent} to construct AAS (Agent-based Learning Simulation) in order to simulate and predict various teaching and learning processes in schools, creating a corresponding knowledge base of teaching experience. This not only has numerous benefits for teacher training and iteration in the education field, but it could also have a profound impact on other industries that require rapid accumulation of experience.

\section{School Simulation}

\subsection{AAS Environment Construction and Design}

\subsubsection{Environment Settings}
\label{sec:environment_setting}

\begin{figure}
    \centering
    \includegraphics[width=1\linewidth]{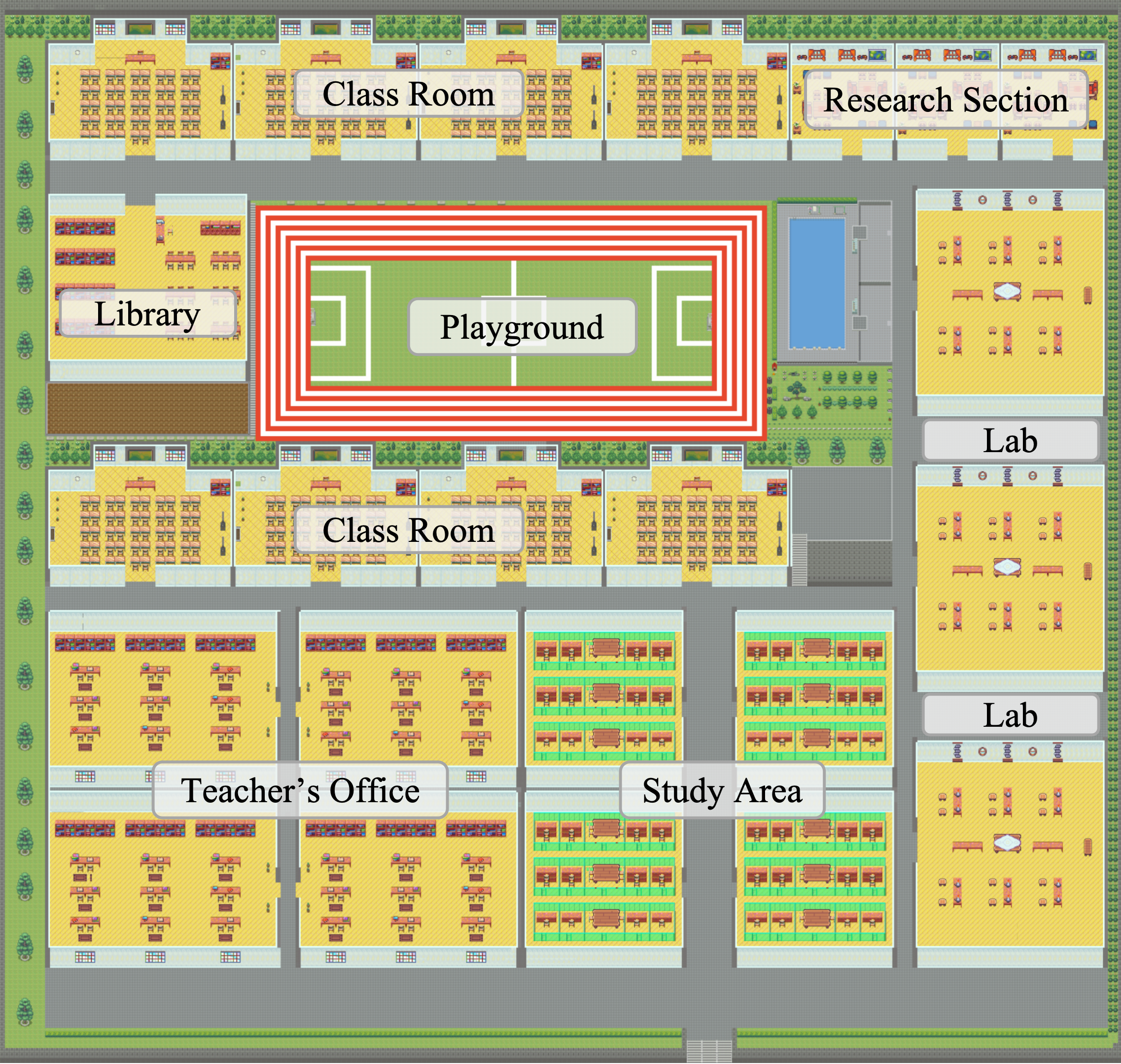}
    \caption{Structural diagram of AAS.}
    \label{fig:Figure1}
\end{figure}

To achieve the visualization of the teaching process, we designed a simulated environment AAS. Inspired by the research of \cite{wang_2023_understanding}, we used Tiled and Cocos to build this environment. Tiled allows for the creation of detailed school layouts, while Cocos serves as the interactive framework for managing the movement and interaction behaviors of the intelligent agents within the school \cite{mohd2023analyzing}. As shown in Figure~\ref{fig:Figure1}, the AAS includes 25 areas, including but not limited to classrooms, libraries, laboratories, and sports fields, providing diverse interactive spaces for both teacher agents and student agents.

\subsubsection{Agent Role Settings}
In the AAS, we designed two main types of interactive roles: teacher agents and student agents. The detailed information and characteristics of these roles were generated using LLM, with QwQ-32B being employed to create rich and diverse role backgrounds and personality traits. The prompts to generate role settings are provided in \ref{appendix:role-settings-generation}.

\subsubsection{Memory Settings}
In the AAS, each agent is equipped with a multi-layered memory system designed to mimic human cognitive processes and manage information beyond the context window of agent. This memory system is structured into three components: Working Memory, and a dual memory base further organized into Short-term and Long-term Memory \cite{fan2024}.
\textbf{Working Memory} corresponds to the context window of the agent, holds currently relevant information that the agent is  processing for decision-making within a short time frame.
Information that exceeds the working memory is stored in a dual memory base \cite{zhao2024expel}, which is fundamentally divided into two types:
\begin{enumerate}
    \item \textbf{Experience Base:} Stores records of past events, interactions, and specific occurrences that the agent has encountered within the simulation. This represents the agent's "lived experiences."
    \item \textbf{Knowledge Base:} Contains structured information related to the agent's role (e.g. academic knowledge for students, teaching methodologies for teachers), general facts, and learned principles. This represents the agent's acquired "knowledge."
\end{enumerate}
The contents of both the experience and knowledge bases are stored and managed within a vector database.
Both the experience base and the knowledge base are further subdivided into \textbf{Short-term and Long-term} components:
\begin{enumerate}
    \item \textbf{Long-term Memory:} Comprises the entirety of the respective Experience Base or Knowledge Base. It serves as a comprehensive repository of all accumulated experiences and knowledge over the agent's simulation lifetime \cite{hochreiter1997long,hatalis2023memory}.
    \item \textbf{Short-term Memory:} Contains a subset of memories from both the Experience and Knowledge bases that the agent deems particularly important or salient at a given time. This selection allows agent to access the most relevant information for current tasks and reflections, mimicking the focus aspect of human short-term memory and attention \cite{hou2024my}.
\end{enumerate}
When retrieving information from the dual memory base, relevance is determined by calculating the cosine similarity between the vector representation of the current query and the vector representations of the memories stored in the database. 
This hierarchical and dual memory structure is fundamental to the Zero-Exp strategy, as it allows agents to retain vast amounts of information beyond the agent's context window, thereby enabling the long-term learning, reflection, and decision-making necessary for autonomous evolution within the dynamic AAS environment.

\subsubsection{Action Settings}

\begin{figure*}[htbp]
    \centering
    \includegraphics[width=1\linewidth]{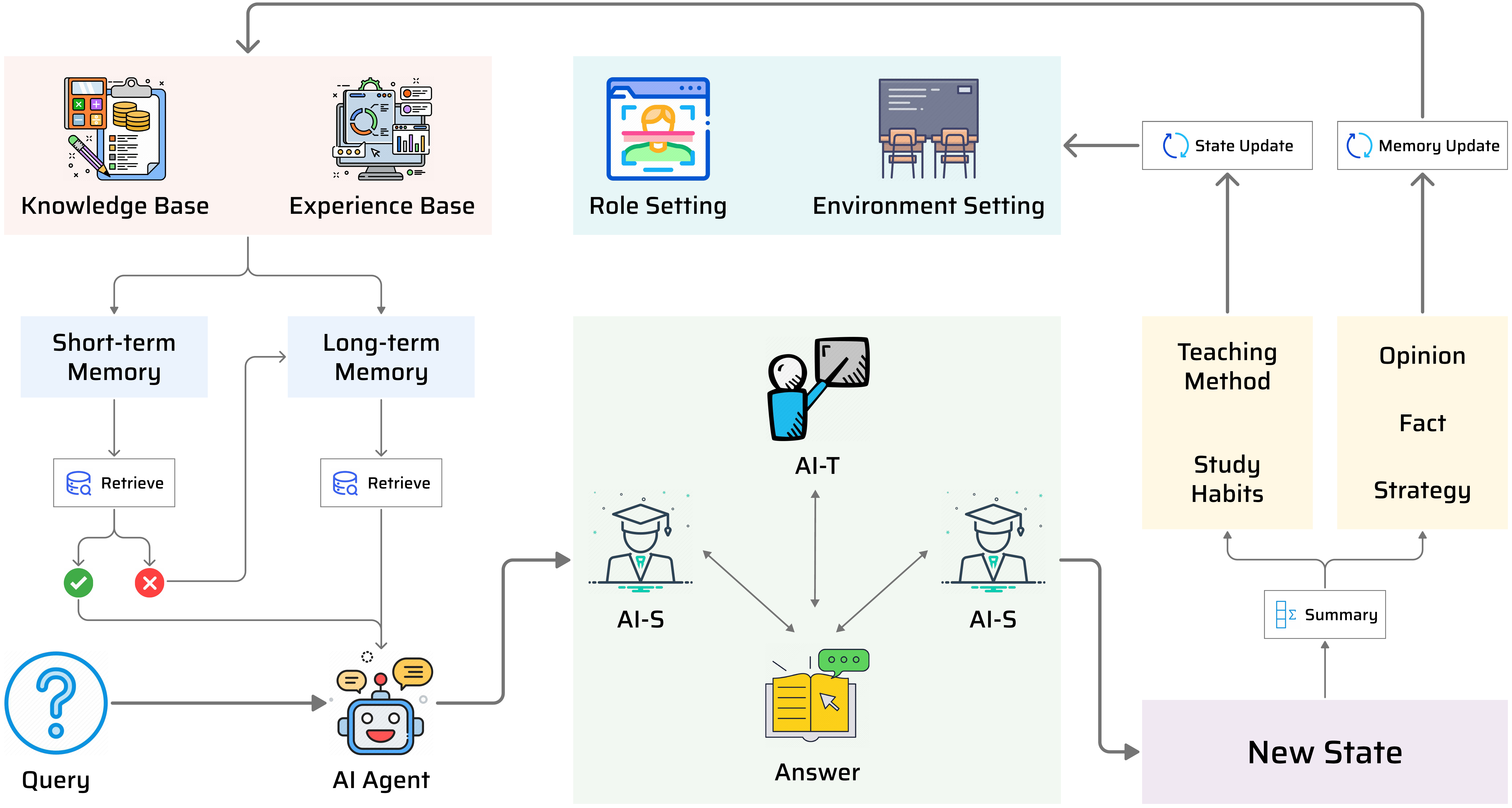}
    \caption{Zero-Exp mechanism}
    \label{fig:zero-exp-mechanism}
\end{figure*}

To simulate educational scenarios, the actions performed by agents within the AAS environment are categorized into distinct sets based on their role: Teacher Agents and Student Agents. These actions allow agents to interact with the environment and other agents, thereby driving the simulation and generating behavioral data  \cite{hu2025exploring,guo2024large}. Agents' action categories and statistics are provided in \ref{appendix:statistics-by-action-category}

\textbf{Teacher Agent Actions:} Teachers primarily perform actions related to teaching, reflection, and managing classroom dynamics \cite{hu2024teaching}. These include {Teaching Practice}, such as conducting lessons, providing guidance; {Teaching Reflection}, involving self-reflection or discussions with other teachers; and {Guidance}, like mediating student disputes and providing social interaction guidance.

\textbf{Student Agent Actions:} Student agents engage in a variety of learning, campus life, and interpersonal activities \cite{zheng2025teaching}. Key actions include {Classroom Learning} , {Laboratory Work}, {Peer Learning/Interaction}, {Self-Directed Learning} , and {Extracurricular Activities}.

This categorization of actions for each agent role, encompassing both positive and negative interactions, facilitates the simulation of realistic educational dynamics, providing structured behavioral data for analysis and agent evolution.

\subsection{Zero-Exp: A Mechanism for Multi-Agent Evolution in AAS}
\label{sec:zero_exp_mechanism}

The Zero-Exp mechanism is central to the self-evolving nature of AAS. It provides a structured process for agents to improve their behaviors based on their simulated experiences and accumulated knowledge \cite{yurtsever2020survey}.

As Figure \ref{fig:zero-exp-mechanism} described, at each step of the simulation, the current state of the AAS environment and the agents' roles are processed. The agent's role settings are defined in the system prompt, ensuring role-consistent behavior. The specifics of the current situation (e.g. location and interaction information) are provided in the final user prompt \cite{xia2024llm}.

To enhance role-playing fidelity \cite{gao2024large}, the retrieval process of the Zero-Exp mechanism is designed to prioritize accessing relevant information from the agent's short-term memory. The mechanism subsequently retrieves relevant information from the long-term memory.

The retrieved memories are then integrated into the final user prompt. The agent's context window is first populated with the final user prompt containing the current situation and retrieved memories, followed by the working memory, which is the previous interaction history. 

Crucially, the agent's response and the subsequent outcomes of its action trigger a process of memory update(\ref{appendix:memory update}) and self-reflection:
\begin{enumerate}
    \item \textbf{Memory Update:} Specific details of the interaction and its results are processed and used to update the agent's Experience Memory Base \cite{sreedhar2025simulating}. New insights, facts, or optimized strategies derived from the interaction or internal reasoning are added to the Knowledge Memory Base. Based on the agent's autonomous selection, selected new memories are also added to the agent's Short-term Memory, ensuring quick access in future relevant situations.
    \item \textbf{State Update:} Agent's reflection or changes in understanding resulting from the interaction and memory updates are also used to dynamically update aspects of their internal Role Setting (e.g., teaching methods, study habits). In some cases, the agent's actions also influence and update the state of the Environment Setting (e.g., moving from a classroom to a teacher's office).
\end{enumerate}

\section{Experiment}
\subsection{Datasets}
\label{sec:datasets}

\begin{figure}
    \centering
    \includegraphics[width=1\linewidth]{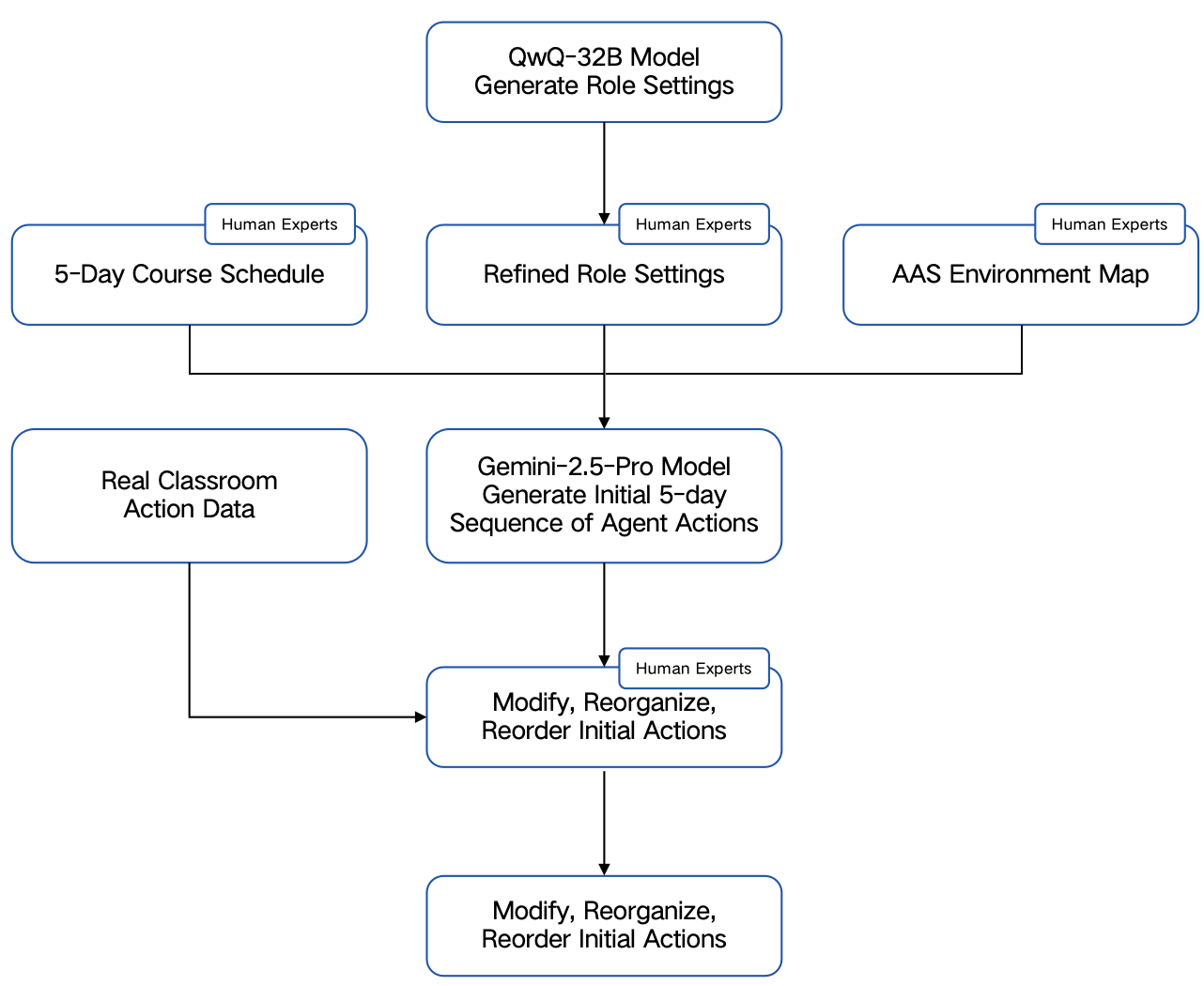}
    \caption{Data building process}
    \label{fig:data-building-process}
\end{figure}

The dataset driving the AAS simulation was constructed through a multi-step process involving LLM generation and expert refinement. This process aimed to create realistic initial conditions and high-fidelity interaction sequences.The overall structure of this dataset generation process is illustrated in Figure~\ref{fig:data-building-process}.

Initial role settings for 10 teacher and 40 student agents were generated using the QwQ-32B model \cite{baker2024simulating}. Experts actively discussed and modified the generated roles while the map(Section  \ref{sec:environment_setting}) and schedule were developed in parallel, ensuring consistency across these foundational elements.
Subsequently, the Gemini-2.5-Pro model generated an initial 5-day sequence of agent actions and interactions based on the refined roles, schedule, and map. This initial data included movements, dialogues, and activity performance \cite{yue2024mathvc}. Specially, classroom teaching action data came from real classrooms.

Finally, this generated data underwent rigorous expert modification, reorganization, and reordering by educational experts to produce the complete and final interaction dataset, serving as the ground truth for evaluation and iteration. This resulting dataset, designated as ID 0: Standard Group, represents plausible and educationally valuable real-world interactions. Specific details of Standard Group are provided in ~\ref{appendix:standard-group-data}.A QA pair in standard group data is as follows:

\begin{center}

\begin{tcolorbox}
[colback=gray!5, colframe=black!80, title=A QA pair, fontupper=\small]
[ \\
    {"role": "system", "content": "Agent's role settings"}, \\
    \\
    // Subsequent turns \\
    {"role": "user", "content": "Time + Environment + Other agent interactions"}, \\
    {"role": "assistant", "content": "Agent action"}, \\
    \\
    // ... more turns ... \\
    \\
    // Current turn's QA \\
    {"role": "user", "content": "Time + Environment + Other agent interactions"}, \\
    {"role": "assistant", "content": "Agent action"}, \\
]
\end{tcolorbox}
\end{center}

\subsection{Experiment Settings}
\label{sec:experimental_settings}

To evaluate the effectiveness of the proposed AAS Zero-Exp mechanism and the contributions of its specific memory components, we used \textbf{GPT-4o, Qwen3-235B-A22B, Qwen3-8B} to act as agents, designed a series of comparative experiments using nine distinct configurations, serving as baselines \cite{gurcan2024llm}. These configurations vary the presence and structure of the external memory base, allowing us to analyze the impact of the dual Experience/Knowledge division and the Short-term/Long-term hierarchy.

The nine experimental configurations are detailed in Table~\ref{tab:experimental_configs}. Each configuration represents a specific setup regarding the agent's access to and organization of external memory, defined by the following parameters:

\textbf{EB (Experience Base)} indicates whether the agent's Experience Memory Base is enabled or disabled.
\textbf{KB (Knowledge Base)} indicates whether the agent's Knowledge Memory Base is enabled or disabled.
\textbf{MB(Memory Base)} describes the structure of the external memory base when enabled: "Dual" signifies separate Experience and Knowledge bases, "Unified" means a single combined base for both, and "None" indicates no external memory base is used.
\textbf{ST/LT (Short-term/Long-term Hierarchy)} indicates whether the Short-term and Long-term memory division and prioritized retrieval mechanism are enabled or disabled within the accessible memory bases.

\begin{table}[ht]
\centering
\small
\renewcommand{\arraystretch}{1.3}
\caption{Experimental settings (Baselines)}
\label{tab:experimental_configs}
\begin{tabular}{lcccc}
\hline
\textbf{ID} & \textbf{EB} & \textbf{KB} & \textbf{MB} & \textbf{ST/LT} \\
\hline
1 & Enabled & Enabled & Dual & Enabled\\
2 & Enabled & Enabled & Unified & Enabled \\
3 & Enabled & Enabled & Dual & Disabled \\
4 & Enabled & Enabled & Unified & Disabled \\
5 & Disabled & Enabled & Dual & Enabled\\
6 & Enabled & Disabled & Dual & Enabled\\
7 & Disabled & Enabled & Dual & Disabled \\
8 & Enabled & Disabled & Dual & Disabled\\
9 & \multicolumn{2}{c}{Disabled} & None & Disabled\\
\hline
\end{tabular}
\end{table}

We conducted simulations for each of these nine configurations using the standard group dataset described in Section \ref{sec:datasets}. The simulation proceeds chronologically, step-by-step, with agent actions and interactions recorded. To ensure that the agent's environment, experience, role setting, and knowledge are appropriately matched with each evaluation point, evaluation is performed periodically during the iterative process.

We employed two primary evaluation methods: an automated metric based on text similarity and a human evaluation based on expert judgment \cite{zhuge2024agent}.

For automated evaluation, we compared the agent's generated response to a reference ground truth answer using the average ROUGE-L \cite{lin2004rouge} scores by every 5\% interval of the total simulation data, reflecting fluency and content overlap. This approach allows us to observe how different memory configurations impact agent performance as they accumulate experience and knowledge throughout the simulated period.






For human evaluation, we selected three configurations from the nine tested: the Full Model (ID 1), the RAG Only (ID 4), and the Context Only (ID 9), along with the original standard group dataset (ID 0). Evaluation was performed at every 10\% data increments throughout the five-day simulation data. At each checkpoint, we extracted one QA pair for every agent (10 teachers and 40 students), resulting in 50 QA pairs per checkpoint and a total of 500 QA pairs for each of the four groups. We recruited nine educational experts to evaluate these QA pairs. For each question, the experts were presented with the four corresponding answers in a blind, randomized order. Without knowing which configuration generated which answer, the experts were asked to vote for the answer they believed best reflected a realistic response in the given educational context. The voting results were then statistically summarized to compare the preference distribution across the four groups.
This blind, head-to-head comparison based on expert opinion provides a valuable qualitative assessment of the simulation fidelity achieved by different memory configurations.



\section{Result}

\subsection{Automated Evaluation Results}
\label{sec:automated-evaluation-results}

\begin{figure}[ht]
    \centering
    \includegraphics[width=1\linewidth]{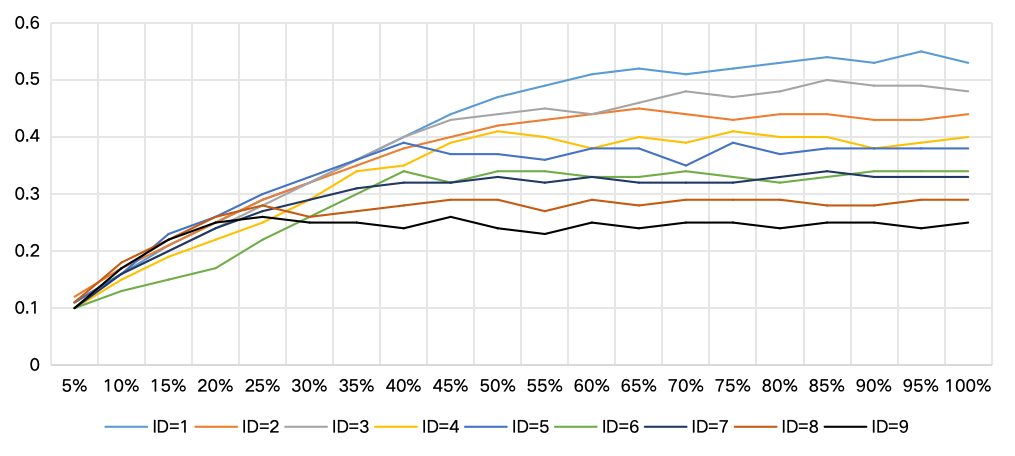}
    \caption{GPT-4o Automated Evaluation Result}
    \label{fig:GPT-4o Automated Evaluation Result}
\end{figure}

\begin{figure}[ht]
    \centering
    \includegraphics[width=1\linewidth]{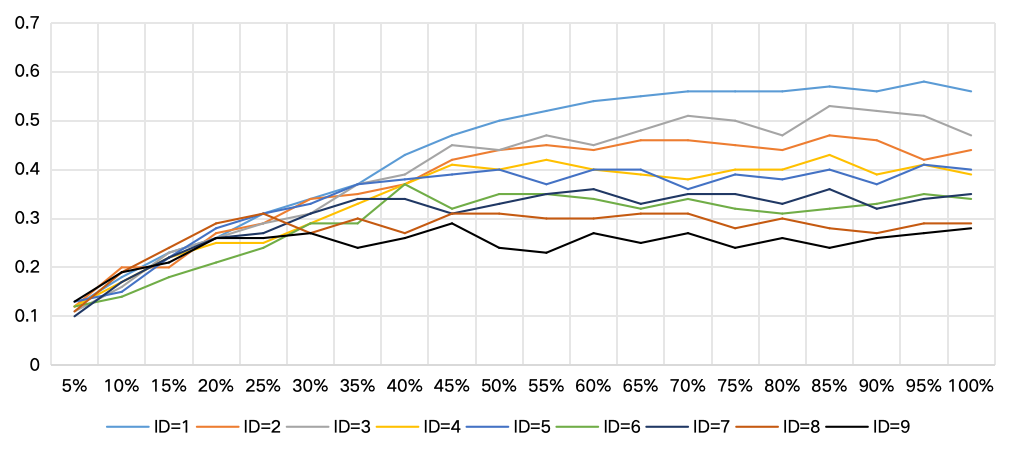}
    \caption{Qwen3-235B-A22B Automated Evaluation Result}
    \label{fig:Qwen3-235B-A22B Automated Evaluation Result}
\end{figure}

\begin{figure}[ht]
    \centering
    \includegraphics[width=1\linewidth]{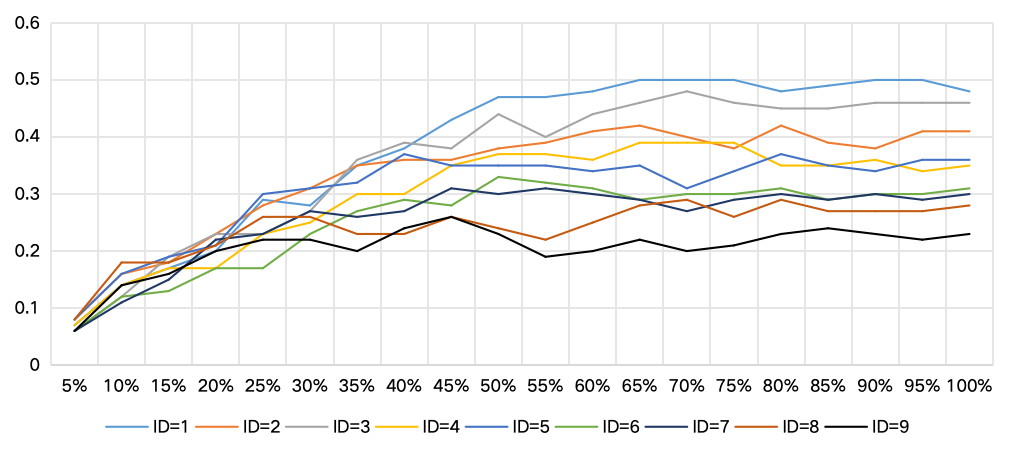}
    \caption{Qwen3-8B Automated Evaluation Result}
    \label{fig:Qwen3-8B Automated Evaluation Result}
\end{figure}

We first analyzed the results of the automated evaluation. Table \ref{tab:gpt-4o automated-evaluation-result}, \ref{tab:qwen3-235B-A22B automated-evaluation-result}, \ref{tab:qwen3-8B automated-evaluation-result} and Figure \ref{fig:GPT-4o Automated Evaluation Result}, \ref{fig:Qwen3-235B-A22B Automated Evaluation Result}, \ref{fig:Qwen3-8B Automated Evaluation Result} presents the average ROUGE-L scores for each configuration at different simulation progress checkpoints.

\begin{table*}[ht]
\renewcommand{\arraystretch}{1.3}
\centering
\small
\caption{GPT-4o Automated Evaluation Result}
\label{tab:gpt-4o automated-evaluation-result}
  \begin{tabular}{cccccccccccc}
    \hline
    \textbf{ID} & \textbf{5\%} & \textbf{10\%} & \textbf{15\%} & \textbf{20\%} & \textbf{25\%} & \textbf{30\%} & \textbf{35\%} & \textbf{40\%} & \textbf{45\%} & \textbf{50\%} \\
    \hline
    1 & 0.11 & 0.16 & 0.21 & 0.25 & 0.29 & 0.32 & 0.36 & 0.40 & 0.44 & 0.47\\
    2 & 0.12 & 0.17 & 0.21 & 0.25 & 0.29 & 0.32 & 0.35 & 0.38 & 0.40 & 0.42\\
    3 & 0.11 & 0.16 & 0.20 & 0.24 & 0.28 & 0.32 & 0.36 & 0.40 & 0.43 & 0.44\\
    4 & 0.10 & 0.15 & 0.19 & 0.22 & 0.25 & 0.29 & 0.34 & 0.35 & \underline{0.39} & 0.41\\
    5 & 0.11 & 0.16 & 0.23 & 0.26 & 0.30 & 0.33 & 0.36 & \underline{0.39} & 0.37 & 0.37\\
    6 & 0.10 & 0.13 & 0.15 & 0.17 & 0.22 & 0.26 & 0.30 & \underline{0.34} & 0.32 & 0.34\\
    7 & 0.10 & 0.16 & 0.20 & 0.24 & 0.27 & 0.29 & 0.31 & \underline{0.32} & 0.32 & 0.33\\
    8 & 0.11 & 0.18 & 0.22 & \underline{0.26} & 0.28 & 0.26 & 0.27 & 0.28 & 0.29 & 0.29\\
    9 & 0.10 & 0.17 & 0.22 & \underline{0.25} & 0.26 & 0.25 & 0.25 & 0.24 & 0.26 & 0.24\\
    \hline

    \textbf{ID} & \textbf{55\%} & \textbf{60\%} & \textbf{65\%} & \textbf{70\%} & \textbf{75\%} & \textbf{80\%} & \textbf{85\%} & \textbf{90\%} & \textbf{95\%} & \textbf{100\%} \\
    \hline
    1 & 0.49 & 0.51 & 0.52 & 0.51 & 0.52 & \underline{0.53} & 0.54 & 0.53 & 0.55 & 0.53\\
    2 & \underline{0.43} & 0.44 & 0.45 & 0.44 & 0.43 & 0.44 & 0.44 & 0.43 & 0.43 & 0.44\\
    3 & 0.45 & 0.44 & 0.46 & \underline{0.48} & 0.47 & 0.48 & 0.48 & 0.50 & 0.49 & 0.49\\
    4 & 0.40 & 0.38 & 0.40 & 0.39 & 0.41 & 0.41 & 0.40 & 0.40 & 0.38 & 0.39\\
    5 & 0.36 & 0.38 & 0.38 & 0.35 & 0.39 & 0.39 & 0.37 & 0.38 & 0.38 & 0.38\\
    6 & 0.34 & 0.33 & 0.33 & 0.34 & 0.33 & 0.32 & 0.33 & 0.34 & 0.34 & 0.34\\
    7 & 0.32 & 0.33 & 0.32 & 0.32 & 0.32 & 0.33 & 0.34 & 0.33 & 0.33 & 0.33\\
    8 & 0.27 & 0.29 & 0.28 & 0.29 & 0.29 & 0.29 & 0.29 & 0.28 & 0.28 & 0.29\\
    9 & 0.23 & 0.25 & 0.24 & 0.25 & 0.25 & 0.25 & 0.25 & 0.24 & 0.25 & 0.25\\
    \hline
  \end{tabular}
\end{table*}

\begin{table*}[ht]
\renewcommand{\arraystretch}{1.3}
\centering
\small
\caption{Qwen3-235B-A22B Automated Evaluation Result}
\label{tab:qwen3-235B-A22B automated-evaluation-result}
  \begin{tabular}{ccccccccccc}
    \hline
    \textbf{ID} & \textbf{5\%} & \textbf{10\%} & \textbf{15\%} & \textbf{20\%} & \textbf{25\%} & \textbf{30\%} & \textbf{35\%} & \textbf{40\%} & \textbf{45\%} & \textbf{50\%} \\
    \hline
    1 & 0.12 & 0.18 & 0.23 & 0.26 & 0.31 & 0.34 & 0.37 & 0.43 & 0.47 & 0.5 \\
    2 & 0.12 & 0.2 & 0.2 & 0.27 & 0.29 & 0.34 & 0.35 & 0.37 & 0.42 & 0.44 \\
    3 & 0.11 & 0.16 & 0.23 & 0.26 & 0.29 & 0.31 & 0.37 & 0.39 & 0.45 & 0.44 \\
    4 & 0.12 & 0.17 & 0.22 & 0.25 & 0.25 & 0.29 & 0.33 & 0.37 & \underline{0.41} & 0.4 \\
    5 & 0.13 & 0.15 & 0.22 & 0.28 & 0.31 & 0.33 & 0.37 & \underline{0.38} & 0.39 & 0.4 \\
    6 & 0.12 & 0.14 & 0.18 & 0.21 & 0.24 & 0.29 & 0.29 & \underline{0.37} & 0.32 & 0.35 \\
    7 & 0.1 & 0.17 & 0.22 & 0.26 & 0.27 & 0.31 & \underline{0.34} & 0.34 & 0.31 & 0.33 \\
    8 & 0.11 & 0.19 & 0.24 & 0.29 & \underline{0.31} & 0.27 & 0.3 & 0.27 & 0.31 & 0.31 \\
    9 & 0.13 & 0.19 & 0.21 & \underline{0.26} & 0.26 & 0.27 & 0.24 & 0.26 & 0.29 & 0.24 \\
    \hline

    \textbf{ID} & \textbf{55\%} & \textbf{60\%} & \textbf{65\%} & \textbf{70\%} & \textbf{75\%} & \textbf{80\%} & \textbf{85\%} & \textbf{90\%} & \textbf{95\%} & \textbf{100\%} \\
    \hline
    1 & 0.52 & 0.54 & 0.55 & \underline{0.56} & 0.56 & 0.56 & 0.57 & 0.56 & 0.58 & 0.56 \\
    2 & \underline{0.45} & 0.44 & 0.46 & 0.46 & 0.45 & 0.44 & 0.47 & 0.46 & 0.42 & 0.44 \\
    3 & 0.47 & 0.45 & 0.48 & \underline{0.51} & 0.5 & 0.47 & 0.53 & 0.52 & 0.51 & 0.47 \\
    4 & 0.42 & 0.4 & 0.39 & 0.38 & 0.4 & 0.4 & 0.43 & 0.39 & 0.41 & 0.39 \\
    5 & 0.37 & 0.4 & 0.4 & 0.36 & 0.39 & 0.38 & 0.4 & 0.37 & 0.41 & 0.4 \\
    6 & 0.35 & 0.34 & 0.32 & 0.34 & 0.32 & 0.31 & 0.32 & 0.33 & 0.35 & 0.34 \\
    7 & 0.35 & 0.36 & 0.33 & 0.35 & 0.35 & 0.33 & 0.36 & 0.32 & 0.34 & 0.35 \\
    8 & 0.3 & 0.3 & 0.31 & 0.31 & 0.28 & 0.3 & 0.28 & 0.27 & 0.29 & 0.29 \\
    9 & 0.23 & 0.27 & 0.25 & 0.27 & 0.24 & 0.26 & 0.24 & 0.26 & 0.27 & 0.28 \\
    \hline
  \end{tabular}
\end{table*}

\begin{table*}[ht]
\renewcommand{\arraystretch}{1.3}
\centering
\small
\caption{Qwen3-8B Automated Evaluation Result}
\label{tab:qwen3-8B automated-evaluation-result}
  \begin{tabular}{ccccccccccc}
    \hline
    \textbf{ID} & \textbf{5\%} & \textbf{10\%} & \textbf{15\%} & \textbf{20\%} & \textbf{25\%} & \textbf{30\%} & \textbf{35\%} & \textbf{40\%} & \textbf{45\%} & \textbf{50\%} \\
    \hline
    1 & 0.07 & 0.14 & 0.17 & 0.2 & 0.29 & 0.28 & 0.35 & 0.38 & 0.43 & 0.47 \\
    2 & 0.08 & 0.16 & 0.18 & 0.23 & 0.28 & 0.31 & 0.35 & 0.36 & 0.36 & 0.38 \\
    3 & 0.06 & 0.12 & 0.19 & 0.23 & 0.23 & 0.27 & 0.36 & 0.39 & 0.38 & 0.44 \\
    4 & 0.07 & 0.14 & 0.17 & 0.17 & 0.23 & 0.25 & 0.3 & 0.3 & 0.35 & \underline{0.37} \\
    5 & 0.08 & 0.16 & 0.19 & 0.21 & 0.3 & 0.31 & 0.32 & \underline{0.37} & 0.35 & 0.35 \\
    6 & 0.06 & 0.12 & 0.13 & 0.17 & 0.17 & 0.23 & 0.27 & 0.29 & 0.28 & \underline{0.33} \\
    7 & 0.06 & 0.11 & 0.15 & 0.22 & 0.23 & 0.27 & 0.26 & 0.27 & \underline{0.31} & 0.3 \\
    8 & 0.08 & 0.18 & 0.18 & 0.21 & \underline{0.26} & 0.26 & 0.23 & 0.23 & 0.26 & 0.24 \\
    9 & 0.06 & 0.14 & 0.16 & 0.2 & \underline{0.22} & 0.22 & 0.2 & 0.24 & 0.26 & 0.23 \\
    \hline

    \textbf{ID} & \textbf{55\%} & \textbf{60\%} & \textbf{65\%} & \textbf{70\%} & \textbf{75\%} & \textbf{80\%} & \textbf{85\%} & \textbf{90\%} & \textbf{95\%} & \textbf{100\%} \\
    \hline
    1 & 0.47 & 0.48 & \underline{0.5} & 0.5 & 0.5 & 0.48 & 0.49 & 0.5 & 0.5 & 0.48 \\
    2 & 0.39 & \underline{0.41} & 0.42 & 0.4 & 0.38 & 0.42 & 0.39 & 0.38 & 0.41 & 0.41 \\
    3 & 0.4 & 0.44 & \underline{0.46} & 0.48 & 0.46 & 0.45 & 0.45 & 0.46 & 0.46 & 0.46 \\
    4 & 0.37 & 0.36 & 0.39 & 0.39 & 0.39 & 0.35 & 0.35 & 0.36 & 0.34 & 0.35 \\
    5 & 0.35 & 0.34 & 0.35 & 0.31 & 0.34 & 0.37 & 0.35 & 0.34 & 0.36 & 0.36 \\
    6 & 0.32 & 0.31 & 0.29 & 0.3 & 0.3 & 0.31 & 0.29 & 0.3 & 0.3 & 0.31 \\
    7 & 0.31 & 0.3 & 0.29 & 0.27 & 0.29 & 0.3 & 0.29 & 0.3 & 0.29 & 0.3 \\
    8 & 0.22 & 0.25 & 0.28 & 0.29 & 0.26 & 0.29 & 0.27 & 0.27 & 0.27 & 0.28 \\
    9 & 0.19 & 0.2 & 0.22 & 0.2 & 0.21 & 0.23 & 0.24 & 0.23 & 0.22 & 0.23 \\
    \hline
  \end{tabular}
\end{table*}

Overall, most configurations show an initial increase in ROUGE-L scores as the agents accumulate experience and knowledge within the simulation, indicating a learning or adaptation process. The performance tends to stabilize or fluctuate in the later stages of the simulation \cite{wei2024editable,liu2023dynamic}.

The full model (ID 1), incorporating both the dual KB/EB structure and the ST/LT hierarchy, achieves the highest ROUGE-L scores, reaching peaks around 0.51-0.55 in the later stages. Comparing ID 1 with configurations that ablate specific memory components allows us to isolate their contributions:

\begin{itemize}
    \item \textbf{Contribution of External Memory:} Comparing ID 9 with any configuration using an external memory base (IDs 1-8) shows a substantial performance gap, demonstrating the fundamental benefit of external memory.
    \item \textbf{Contribution of Dual KB/EB Structure:} Comparing configurations with similar ST/LT structures but different base organizations reveals the advantage of the dual structure. ID 1 consistently outperforms ID 2. Similarly, ID 3 generally performs better than ID 4, although the gap is smaller in some phases. This suggests that maintaining separate repositories for experience and knowledge is beneficial for more effective retrieval and utilization.
    \item \textbf{Contribution of ST/LT Hierarchy:} Comparing configurations with similar base structures but different ST/LT organizations shows the benefit of the short-term memory mechanism. ID 1 outperforms ID 3. ID 2 outperforms ID 4. Furthermore, comparing single-base configurations like ID 5  vs. ID 7  and ID 6 vs. ID 8 consistently shows the advantage of incorporating the ST/LT hierarchy. This indicates that prioritizing recently salient memories in Short-term memory significantly enhances the agent's ability to generate relevant responses.
\end{itemize}

In summary, the automated evaluation results strongly support the effectiveness of the proposed Zero-Exp mechanism's memory structure. The full model (ID 1) achieves the highest ROUGE-L scores, demonstrating that the combination of a dual experience/knowledge base and a hierarchical Short-term/Long-term memory organization significantly enhances agent performance in generating responses aligned with the reference data throughout the simulation.

\subsection{Human Evaluation Results}
\label{sec:human-evaluation-results}

\begin{figure}[ht]
    \centering
    \includegraphics[width=1\linewidth]{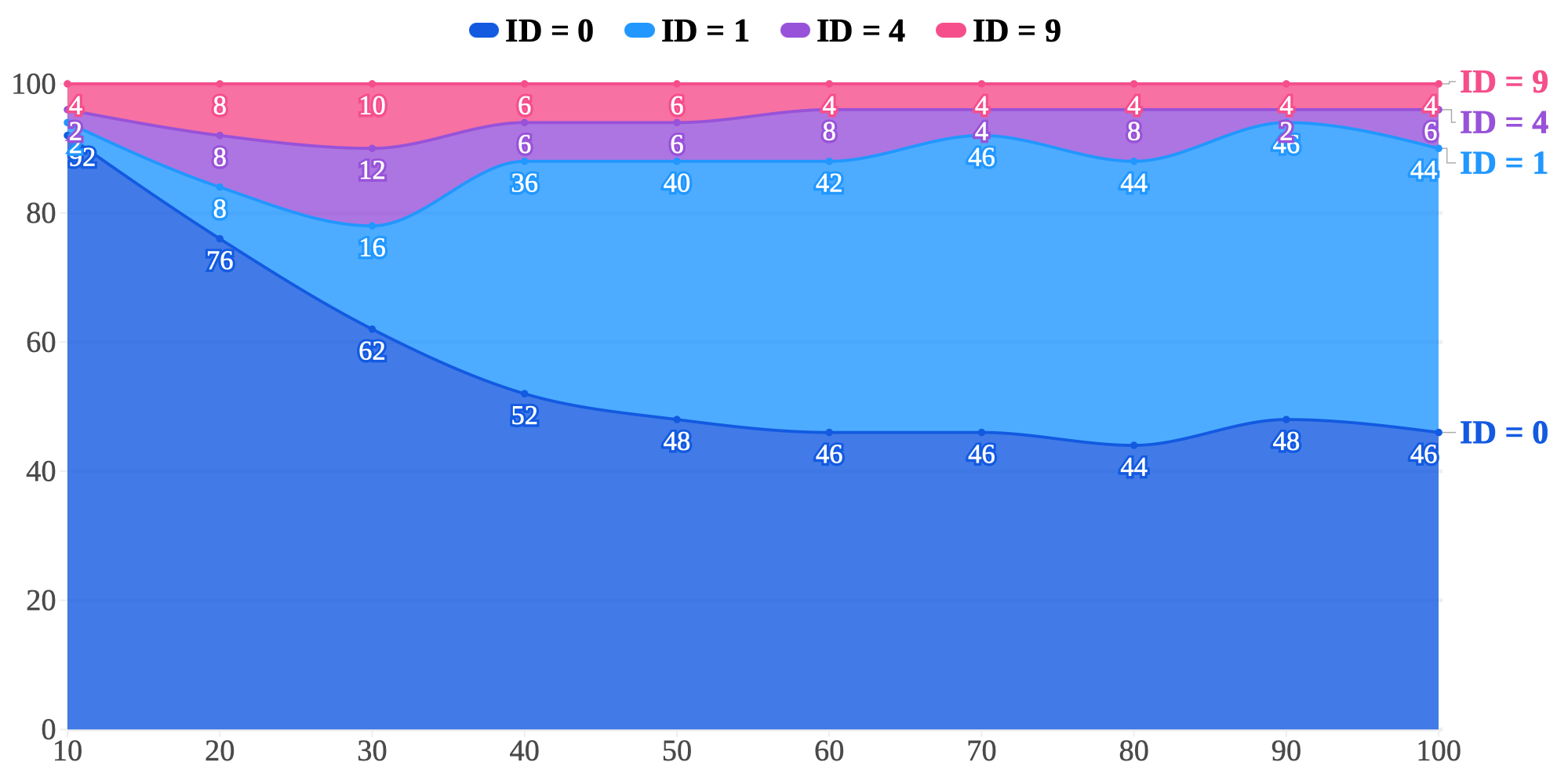}
    \caption{Human Evaluation Result}
    \label{fig:Human Evaluation Result}
\end{figure}

We also conducted a human evaluation involving educational experts to assess the perceived realism and quality of agent(acted by GPT-4o) interactions \cite{samuel2024personagym}. Table \ref{tab:huaman-evaluation-result} and Figure \ref{fig:Human Evaluation Result} presents the percentage of QA pairs (out of 50 per checkpoint) for which each group's answer was voted as the most realistic by the experts at different simulation progress checkpoints.

The results show significant differences in perceived realism. The baselines without the full memory structure, like Context Only (ID 9) and Unified Memory (LT Only) (ID 4), received low preference votes throughout the simulation, highlighting the necessity of a comprehensive memory system for generating realistic interactions.

\begin{table}[ht]
\centering
\small
\renewcommand{\arraystretch}{1.3}
\caption{Human Evaluation Result}
\label{tab:huaman-evaluation-result}
\begin{tabular}{lccccc}
\hline
\textbf{ID} & \textbf{10\%} & \textbf{20\%} & \textbf{30\%} & \textbf{40\%} & \textbf{50\%} \\
\hline
0 & 92\% & 76\% & 62\% & 52\% & 48\% \\
1 & 2\% & 8\% & 16\% & 36\% & 40\% \\
4 & 2\% & 8\% & 12\% & 6\% & 6\% \\
9 & 4\% & 8\% & 10\% & 6\% & 6\% \\
\hline
\textbf{ID} & \textbf{60\%} & \textbf{70\%} & \textbf{80\%} & \textbf{90\%} & \textbf{100\%} \\
\hline
0 & 46\% & 46\% & 44\% & 48\% & 46\% \\
1 & 42\% & 46\% & 44\% & 46\% & 44\% \\
4 & 8\% & 4\% & 8\% & 2\% & 6\% \\
9 & 4\% & 4\% & 4\% & 4\% & 4\% \\
\hline
\end{tabular}
\end{table}

In contrast, the full model (ID 1) demonstrates a strong learning curve. Starting with low preference in the early stages, its performance rapidly increases as the simulation progresses. Notably, the full model's perceived realism approaches and stabilizes near that of the standard group (ID 0) in the later stages (from 60\%). While the standard group represents the expert-curated ground truth and shows high preference initially, the evolving agents in the full model configuration generate interactions that experts perceive as comparably realistic over time.

In conclusion, the human evaluation results corroborate the findings from the automated evaluation. The full model (ID 1) significantly outperforms the baselines (ID 4 and ID 9) in generating interactions deemed realistic by experts. The convergence of ID 1's performance with the Reference Group (ID 0) in the later stages provides strong qualitative evidence that the Zero-Exp mechanism, powered by the proposed memory system, enables multi-agent evolution towards generating high-fidelity educational simulations.

\section{Conclusion}

We introduces the AAS, a multi-agent simulation environment designed to model and accelerate the evolution of educational cognitive processes through situated interactions. Addressing lack of systematic teaching process modeling and challenges in simulating diverse participant behaviors, we proposed the Zero-Exp mechanism. Central to Zero-Exp is a dual memory base, distinguishing between episodic experience and structured knowledge, organized hierarchically into Short-term and Long-term components. This architecture facilitates a continuous "experience-reflection-optimization" cycle, enabling agents to evolve autonomously based on their interactions within the simulated school environment.

Our comprehensive experimental evaluation, involving nine different memory configurations, validates the effectiveness of the Zero-Exp mechanism. Both automated ROUGE-L scores and expert human evaluation demonstrate that the full memory model (ID 1) significantly outperforms baselines lacking the dual structure or the ST/LT hierarchy. The results indicate that the proposed memory system is crucial for enabling agents to generate more realistic appropriate behaviors, progressively aligning with expert-curated ground truth data over time.

The AAS environment and Zero-Exp mechanism represent a significant step towards creating high-fidelity digital twins of educational settings and generating valuable behavioral data \cite{vsturm2024enhancing}. This work provides a verifiable technical model and theoretical pathway for future research in educational AI, agent-based simulation, and the broader pursuit of experience-driven artificial intelligence, contributing foundational elements towards realizing the potential of the "Era of Experience" in educational and potentially other complex human-centric domains.

\section{Limitations}

Despite the promising results, our work has several limitations. Firstly, the current simulation scale is relatively limited, involving only 50 agents (10 teachers and 40 students) over a 5-day period. Scaling to a full school environment with hundreds or thousands of agents and longer durations presents significant computational and design challenges. Secondly, the agents' cognitive abilities are primarily based on LLMs. While powerful for text-based reasoning and interaction, the absence of Vision-Language Models (VLMs) means agents lack the ability to visually perceive and interpret their environment or the non-verbal cues of other agents, limiting the realism of situated interactions that rely on visual context. Thirdly, the fidelity of the simulation is also dependent on the quality and diversity of the initial expert-curated dataset used for evaluation and guiding the initial evolution. While refined, it represents a specific set of scenarios. Furthermore, the vast complexity of human cognition, social interaction, and the full spectrum of teaching and learning processes in real educational settings are difficult to fully capture, and while AAS makes significant strides, there are still nuances that may not be perfectly replicated. Finally, the reliance on LLMs means the agents' behaviors are inherently constrained by the capabilities and potential biases of the underlying models.

Future work should focus on scaling the AAS environment to accommodate larger numbers of agents and more complex scenarios. Incorporating VLMs or other multimodal models could enhance agents' perception and interaction capabilities by allowing them to process visual information. Exploring alternative or hybrid LLM architectures could further enhance agent reasoning and interaction capabilities. Developing more sophisticated reflection and optimization mechanisms within the Zero-Exp framework could accelerate and refine agent evolution. Applying the generated high-fidelity data to specific educational applications, such as personalized learning pathway design or automated feedback systems, is a crucial next step. Finally, extending the Zero-Exp mechanism and the AAS framework to other domains requiring the accumulation and utilization of complex experience could demonstrate the broader applicability of this approach.

\section{Ethical Considerations}

This study was conducted with full consideration of ethical principles and adherence to research standards. We recruited participants from higher education institutions in China, specifically targeting university teachers as our primary participants as educational experts. All participants were provided with comprehensive written informed consent forms that detailed the purpose and scope of the research, data collection and usage protocols, potential risks and benefits of participation, their right to withdraw from the study at any time, and contact information for the research team. Participants were compensated fairly for their time and contribution, with payment rates determined based on standard academic research compensation in China. The compensation was deemed appropriate considering the participants' professional status and local economic conditions. The research protocol, including all data collection methods and informed consent procedures, was reviewed and approved by  Ethics Review Board. All participants were informed about how their input would contribute to the development and refinement of the AI-Agent School system.

\clearpage

\appendix
\renewcommand{\thesection}{Appendix~\Alph{section}}
\renewcommand{\thesubsection}{\thesection.\arabic{subsection}}

\section{Role Settings Generation}
\label{appendix:role-settings-generation}

This section provides prompts used to generate the initial role settings for the teacher and student agents using the QwQ-32B model. The prompt guided the model to create diverse and detailed profiles, including personality traits, backgrounds, and specific characteristics relevant to their roles within the AI-Agent School simulation.

\begin{center}
\begin{tcolorbox}[colback=gray!5,colframe=black!80,title=Teacher Agent, fontupper=\small]
You are a creative writer tasked with generating a detailed profile for an agent in an educational simulation. Create a unique and realistic profile based on the specified role.\\
Generate a profile for a middle school Chinese/Math/Physics/Chemistry/History teacher. Include:\\
- Full Name \\
- Gender \\
- Age \\
- Years of teaching experience \\
- Teaching Philosophy/Style (e.g., strict, supportive, innovative, traditional) \\
- Personality Traits (e.g., patient,  enthusiastic, strict, humorous, introverted, extroverted) \\
- Strengths as a teacher \\
- Weaknesses as a teacher \\
- Interests or hobbies outside of teaching \\
- Any specific quirks or habits \\
Ensure the generated profile is internally consistent and provides enough detail to inform realistic behavior within a school simulation environment.
\end{tcolorbox}
\end{center}

\begin{center}
\begin{tcolorbox}[colback=gray!5,colframe=black!80,title=Stduent Agent, fontupper=\small]
You are a creative writer tasked with generating a detailed profile for an agent in an educational simulation. Create a unique and realistic profile based on the specified role.\\
Generate a profile for a middle school student. Include: \\
- Full Name \\
- Academic Performance (e.g., excellent, average, struggling) \\
- Learning Style (e.g., visual, auditory, kinesthetic, independent, collaborative) \\
- Personality Traits (e.g., shy, outgoing, curious, diligent, easily distracted, rebellious) \\
- Brief Background Story (e.g., family background, significant life events, motivation for learning) \\
- Academic Strengths \\
- Academic Weaknesses \\
- Interests or hobbies outside of school
- Social Tendencies (e.g., popular, quiet, leader, follower) \\
- Any specific quirks or habits \\
Ensure the generated profile is internally consistent and provides enough detail to inform realistic behavior within a school simulation environment.
\end{tcolorbox}
\end{center}

\section{Memory Update}
\label{appendix:memory update}

This section provides prompt to updating experience base and knowledge base with long-term and short-term memory.

\begin{center}
\begin{tcolorbox}[colback=gray!5,colframe=black!80,title=Memory Update, fontupper=\small]
You are an AI agent in a school simulation. Your task is to process recent events to update and refine your four memory components: Long-term Experience, Short-term Experience, Long-term Knowledge, and Short-term Knowledge. \\
Current Situation: [Current environment, ongoing activity, and recent interaction details] \\
Recent Experience: [Detailed log of the agent's actions, observations, and interactions in the immediate past] \\
Based on the Current Situation, Recent Experience, perform the following steps to update your memory: \\
1. Analyze the Recent Experience: Identify the key events, interactions, and observations that occurred in the immediate past. \\
2.  Integrate with Retrieved Memories: Compare and contrast the Recent Experience with the Current Situation. \\
3. Identify New Information and Refinements: Extract any new facts, insights, specific event details, or observations from the Recent Experience and its integration with past memories. Also, identify any existing entries in your Long-term Memory Bases that should be updated, corrected, or reinforced based on this new information. \\
4. Formulate Updates for Long-term Experience Memory: Based on the analysis in steps 1-3, generate the specific content to be added to or update your Long-term Experience Memory Base. \\
5. Formulate Updates for Long-term Knowledge Memory: Based on the analysis in steps 1-3, generate the specific content to be added to or update your Long-term Knowledge Memory Base. \\
6. Select Salient Information for Short-term Memory: From the combination of the Recent Experience and the updates formulated for your Long-term Memory Bases (steps 4 and 5), identify the most currently important or salient pieces of information. These are items that are highly relevant to the current context and potential near-future situations, and should be prioritized for quick access in your Short-term Memory. \\
7. Formulate Content for Short-term Experience Memory: Based on step 6, generate the specific content related to experiences to be added to your Short-term Experience Memory. \\
8. Formulate Content for Short-term Knowledge Memory: Based on step 6, generate the specific content related to knowledge to be added to your Short-term Knowledge Memory. \\
Output the results as a JSON object with the following structure: \\
\texttt{\{ }\\
long\_term\_experience\_updates: [string], \\
long\_term\_knowledge\_updates: [string], \\
short\_term\_experience\_content: [string], \\
short\_term\_knowledge\_content: [string] \\
\texttt{\} }
\end{tcolorbox}
\end{center}
\section{Role Settings Update}
\label{appendix:role-settings-update}
This section provides prompt to update Agent Role Setting based on accumulated experience and reflection.
\begin{center}
\begin{tcolorbox}[colback=gray!5,colframe=black!80,title=Role Settings Update, fontupper=\small]
You are an AI agent reflecting on your recent experiences and learning. Your task is to update in your current role setting within the school simulation. \\
Current Role Setting: [] \\
Reflection Insights Summary: [Key learnings, insights, and salient information derived from the recent Memory Summary process.] \\
Based on your Current Role Setting and the Reflection Insights Summary, identify which aspects of your profile should be updated within the simulated environment. Consider areas like: \\
- Personality Traits (e.g., becoming more patient, less easily distracted, more proactive)\\
- Behavioral Tendencies (e.g., more likely to ask questions, more likely to collaborate, changing reaction patterns to certain stimuli)\\
- Strategies (e.g., adjusting teaching methods, changing study habits, modifying interaction approaches)\\
- Beliefs or Understandings related to your role and the simulation environment\\
- Specific Quirks or Habits \\
Describe the proposed updates to your role setting based on your analysis. If no significant updates are deemed necessary based on the recent reflections, state that the role setting remains largely unchanged.
\end{tcolorbox}
\end{center}
\section{Standard Group Data}
\label{appendix:standard-group-data}

\subsection{Class Timetables}
\label{appendix:class-timetables}

\begin{table*}[b]
\caption{Class 1 Timetable}
\centering
\begin{tabular}{l|ccccc}
\hline
\textbf{Time Slot} & \textbf{Monday}  & \textbf{Tuesday} & \textbf{Wednesday} & \textbf{Thursday} & \textbf{Friday} \\
\hline
Period 1 (8:00-8:40) & Chinese & Math & Physics & Chemistry  & History \\
Break 1  & - & - & - & - & - \\
Period 2 (9:00-9:40) & Math & History & Chemistry & Chinese & Physics \\
Break 2 & - & - & - & - & - \\
Period 3 (10:00-10:40) & Physics & Chinese & Math & History & Chemistry \\
Lunch Break & - & - & - & - & - \\
Period 4 (13:30-14:10) & Chemistry & Physics & History & Math & Chinese \\
Break 3 & - & - & -  & - & - \\
Period 5 (14:30-15:10) & History & Chemistry & Chinese & Physics & Math \\

Extracurricular Activity & - & - & - & - & - \\
\hline
\end{tabular}
\end{table*}

\begin{table*}[b]
\centering
\caption{Class 2 Timetable}
\begin{tabular}{l|ccccc}
\hline
\textbf{Time Slot} & \textbf{Monday}  & \textbf{Tuesday} & \textbf{Wednesday} & \textbf{Thursday} & \textbf{Friday} \\
\hline 
Period 1 (8:00-8:40) & Math & Chemistry & Chinese & History & Physics \\
Break 1 & - & - & - & - & - \\
Period 2 (9:00-9:40) & Physics & Math & Physics & Chemistry & Chinese \\
Break 2 & - & - & - & - & - \\
Period 3 (10:00-10:40) & Chemistry & History & History & Chinese & Math \\
Lunch Break & - & - & - & - & - \\
Period 4 (13:30-14:10) & History & Chinese & Chemistry & Physics & Chemistry \\
Break 3 & - & - & - & - & - \\
Period 5 (14:30-15:10) & Chinese & Physics & Math    & Math & History \\
Extracurricular Activity & - & - & - & - & - \\
\hline
\end{tabular}
\end{table*}

\subsection{Statistics by Action Category}
\label{appendix:statistics-by-action-category}

\begin{table}[h]
\centering
\caption{Statistics by Action Category}
\begin{tabular}{ccc}
\hline
 \textbf{Role} & \textbf{Action} & \textbf{Counts} \\
 \hline
 Teacher & Teaching Practice & 12873 \\
 Teacher & Teaching Reflection & 795 \\
 Teacher & Other Guidance & 412 \\
 Student & Classroom Learning & 681 \\
 Student & Laboratory Work & 2476 \\
 Student & Peer Learning/Interaction & 9499 \\
 Student & Self-Directed Learning & 1708 \\
 Student & Extracurricular Activities & 8532 \\
 \hline
 
\end{tabular}
\end{table}

\bibliography{custom}

\begin{thebibliography}{53}
\providecommand{\natexlab}[1]{#1}

\bibitem[{Atkinson(2002)}]{atkinson_2002_optimizing}
Robert~K. Atkinson. 2002.
\newblock \href {https://doi.org/10.1037/0022-0663.94.2.416} {Optimizing
  learning from examples using animated pedagogical agents.}
\newblock \emph{Journal of Educational Psychology}, 94:416--427.

\bibitem[{Baker and Azher(2024)}]{baker2024simulating}
Zachary~R Baker and Zarif~L Azher. 2024.
\newblock Simulating the us senate: An llm-driven agent approach to modeling
  legislative behavior and bipartisanship.
\newblock \emph{arXiv preprint arXiv:2406.18702}.

\bibitem[{Baylor(1999)}]{baylor_1999_intelligent}
Amy Baylor. 1999.
\newblock Intelligent agents as cognitive tools for education.
\newblock \emph{Educational Technology archive}, 39:36--40.

\bibitem[{Baylor and Kim(2005)}]{baylor_2005_simulating}
Amy~L Baylor and Yanghee Kim. 2005.
\newblock Simulating instructional roles through pedagogical agents.
\newblock \emph{International Journal of Artificial Intelligence in Education},
  15(2):95--115.

\bibitem[{Brusilovsky et~al.(2003)Brusilovsky, Corbett, and
  de~Rosis}]{goodman_2003_towards}
Peter Brusilovsky, Albert Corbett, and Fiorella de~Rosis, editors. 2003.
\newblock \emph{Towards intelligent agents for collaborative learning:
  Recognizing the roles of dialogue participants}. Springer Berlin Heidelberg.

\bibitem[{Cassell(2001)}]{cassell_2001_embodied}
Justine Cassell. 2001.
\newblock \href {https://doi.org/10.1609/aimag.v22i4.1593} {Embodied
  conversational agents: representation and intelligence in user interfaces}.
\newblock \emph{Ai Magazine}, 22:67--83.

\bibitem[{Chen et~al.(2024)Chen, Hu, and Wang}]{chen_2024_empowering}
Xiaojiao Chen, Zhebing Hu, and Chengliang Wang. 2024.
\newblock \href {https://doi.org/10.1007/s10639-024-12549-7} {Empowering
  education development through aigc: A systematic literature review}.
\newblock \emph{Education and Information Technologies}.

\bibitem[{Dai et~al.(2024)Dai, Ke, Pan, Moon, and Liu}]{dai_2024_effects}
Chih-Pu Dai, Fengfeng Ke, Yanjun Pan, Jewoong Moon, and Zhichun Liu. 2024.
\newblock \href {https://doi.org/10.1007/s10648-024-09855-4} {Effects of
  artificial intelligence-powered virtual agents on learning outcomes in
  computer-based simulations: A meta-analysis}.
\newblock \emph{Educational psychology review}, 36.

\bibitem[{Fan et~al.(2024)Fan, Ding, Ning, Wang, Li, Yin, Chua, and
  Li}]{fan2024}
Wenqi Fan, Yujuan Ding, Liangbo Ning, Shijie Wang, Hengyun Li, Dawei Yin,
  Tat-Seng Chua, and Qing Li. 2024.
\newblock \href {https://doi.org/10.1145/3637528.3671470} {A survey on rag
  meeting llms: Towards retrieval-augmented large language models}.
\newblock In \emph{Proceedings of the 30th ACM SIGKDD Conference on Knowledge
  Discovery and Data Mining}, KDD '24, page 6491–6501, New York, NY, USA.
  Association for Computing Machinery.

\bibitem[{Gao et~al.(2024)Gao, Lan, Li, Yuan, Ding, Zhou, Xu, and
  Li}]{gao2024large}
Chen Gao, Xiaochong Lan, Nian Li, Yuan Yuan, Jingtao Ding, Zhilun Zhou, Fengli
  Xu, and Yong Li. 2024.
\newblock Large language models empowered agent-based modeling and simulation:
  A survey and perspectives.
\newblock \emph{Humanities and Social Sciences Communications}, 11(1):1--24.

\bibitem[{Guo et~al.(2024)Guo, Chen, Wang, Chang, Pei, Chawla, Wiest, and
  Zhang}]{guo2024large}
Taicheng Guo, Xiuying Chen, Yaqi Wang, Ruidi Chang, Shichao Pei, Nitesh~V
  Chawla, Olaf Wiest, and Xiangliang Zhang. 2024.
\newblock Large language model based multi-agents: A survey of progress and
  challenges.
\newblock \emph{arXiv preprint arXiv:2402.01680}.

\bibitem[{G{\"u}rcan(2024)}]{gurcan2024llm}
{\"O}nder G{\"u}rcan. 2024.
\newblock Llm-augmented agent-based modelling for social simulations:
  Challenges and opportunities.
\newblock \emph{HHAI 2024: Hybrid Human AI Systems for the Social Good}, pages
  134--144.

\bibitem[{Hatalis et~al.(2023)Hatalis, Christou, Myers, Jones, Lambert,
  Amos-Binks, Dannenhauer, and Dannenhauer}]{hatalis2023memory}
Kostas Hatalis, Despina Christou, Joshua Myers, Steven Jones, Keith Lambert,
  Adam Amos-Binks, Zohreh Dannenhauer, and Dustin Dannenhauer. 2023.
\newblock Memory matters: The need to improve long-term memory in llm-agents.
\newblock In \emph{Proceedings of the AAAI Symposium Series}, volume~2, pages
  277--280.

\bibitem[{Hayes-Roth(1995)}]{hayesroth_1995_an}
Barbara Hayes-Roth. 1995.
\newblock \href {https://doi.org/10.1016/0004-3702(94)00004-k} {An architecture
  for adaptive intelligent systems}.
\newblock \emph{Artificial Intelligence}, 72:329--365.

\bibitem[{Hochreiter and Schmidhuber(1997)}]{hochreiter1997long}
Sepp Hochreiter and J{\"u}rgen Schmidhuber. 1997.
\newblock Long short-term memory.
\newblock \emph{Neural computation}, 9(8):1735--1780.

\bibitem[{Hou et~al.(2024)Hou, Tamoto, and Miyashita}]{hou2024my}
Yuki Hou, Haruki Tamoto, and Homei Miyashita. 2024.
\newblock " my agent understands me better": Integrating dynamic human-like
  memory recall and consolidation in llm-based agents.
\newblock In \emph{Extended Abstracts of the CHI Conference on Human Factors in
  Computing Systems}, pages 1--7.

\bibitem[{Hu et~al.(2024)Hu, Zheng, Zhu, Ding, Wang, and Gu}]{hu2024teaching}
Bihao Hu, Longwei Zheng, Jiayi Zhu, Lishan Ding, Yilei Wang, and Xiaoqing Gu.
  2024.
\newblock Teaching plan generation and evaluation with gpt-4: Unleashing the
  potential of llm in instructional design.
\newblock \emph{IEEE Transactions on Learning Technologies}.

\bibitem[{Hu et~al.(2025)Hu, Zhu, Pei, and Gu}]{hu2025exploring}
Bihao Hu, Jiayi Zhu, Yiying Pei, and Xiaoqing Gu. 2025.
\newblock Exploring the potential of llm to enhance teaching plans through
  teaching simulation.
\newblock \emph{npj Science of Learning}, 10(1):7.

\bibitem[{Jin et~al.(2024)Jin, Lee, Shin, and Kim}]{jin_2024_teach}
Hyoungwook Jin, Seonghee Lee, Hyungyu Shin, and Juho Kim. 2024.
\newblock \href {https://doi.org/10.1145/3613904.3642349} {Teach ai how to
  code: Using large language models as teachable agents for programming
  education}.
\newblock \emph{In Proceedings of the CHI Conference on Human Factors in
  Computing Systems}.

\bibitem[{Jing et~al.(2024)Jing, Wang, Chen, and Wang}]{jing_2024_what}
Yuhui Jing, Haoming Wang, Xiaojiao Chen, and Chengliang Wang. 2024.
\newblock \href {https://doi.org/10.1057/s41599-024-02751-w} {What factors will
  affect the effectiveness of using chatgpt to solve programming problems? a
  quasi-experimental study}.
\newblock \emph{Humanities and Social Sciences Communications}, 11:1–12.

\bibitem[{Jing et~al.(2023)Jing, Zhao, Zhu, Wang, Wang, and
  Xia}]{jing_2023_research}
Yuhui Jing, Leying Zhao, Keke Zhu, Haoming Wang, Chengliang Wang, and Qi~Xia.
  2023.
\newblock \href {https://doi.org/10.3390/su15043115} {Research landscape of
  adaptive learning in education: A bibliometric study on research publications
  from 2000 to 2022}.
\newblock \emph{Sustainability}, 15:3115--3115.

\bibitem[{Jinxin et~al.(2023)Jinxin, Jiabao, Yilei, Xingjiao, Jiawen, and
  Liang}]{jinxin_2023_cgmi}
Shi Jinxin, Zhao Jiabao, Wang Yilei, Wu~Xingjiao, Li~Jiawen, and He~Liang.
  2023.
\newblock \href {https://arxiv.org/abs/2308.12503} {Cgmi: Configurable general
  multi-agent interaction framework}.

\bibitem[{Kim and Baylor(2015)}]{kim_2015_researchbased}
Yanghee Kim and Amy~L. Baylor. 2015.
\newblock \href {https://doi.org/10.1007/s40593-015-0055-y} {Research-based
  design of pedagogical agent roles: a review, progress, and recommendations}.
\newblock \emph{International Journal of Artificial Intelligence in Education},
  26:160--169.

\bibitem[{Kim and Lim(2013)}]{kim_2013_gendered}
Yanghee Kim and Jae~Hoon Lim. 2013.
\newblock \href {https://doi.org/10.1037/a0031027} {Gendered socialization with
  an embodied agent: Creating a social and affable mathematics learning
  environment for middle-grade females.}
\newblock \emph{Journal of Educational Psychology}, 105:1164--1174.

\bibitem[{Kommers and Richards(2005)}]{veletsianos_2005_the}
Piet Kommers and Griff Richards, editors. 2005.
\newblock \href {https://www.learntechlib.org/p/20645} {\emph{The role of
  intelligent agents on learner performance}}. Association for the Advancement
  of Computing in Education. Association for the Advancement of Computing in
  Education (AACE).

\bibitem[{Lan and Chen(2024)}]{lan_2024_teachers}
Yu-Ju Lan and Nian-Shing Chen. 2024.
\newblock \href {https://www.jstor.org/stable/48754837} {Teachers’ agency in
  the era of llm and generative ai: Designing pedagogical ai agents}.
\newblock \emph{Educational Technology \& Society}, 27(1):pp. I–XVIII.

\bibitem[{Li et~al.(2024)Li, Wang, Zhang, Li, Lai, Kang, Ma, and
  Liu}]{li_2024_agent}
Junkai Li, Siyu Wang, Meng Zhang, Weitao Li, Yunghwei Lai, Xinhui Kang, Weizhi
  Ma, and Yang Liu. 2024.
\newblock \href {https://doi.org/10.48550/arXiv.2405.02957} {Agent hospital: A
  simulacrum of hospital with evolvable medical agents}.

\bibitem[{Lin(2004)}]{lin2004rouge}
Chin-Yew Lin. 2004.
\newblock Rouge: A package for automatic evaluation of summaries.
\newblock In \emph{Text summarization branches out}, pages 74--81.

\bibitem[{Liu et~al.(2023)Liu, Zhang, Li, Liu, and Yang}]{liu2023dynamic}
Zijun Liu, Yanzhe Zhang, Peng Li, Yang Liu, and Diyi Yang. 2023.
\newblock Dynamic llm-agent network: An llm-agent collaboration framework with
  agent team optimization.
\newblock \emph{arXiv preprint arXiv:2310.02170}.

\bibitem[{Madni and Madni(2008)}]{madni_2008_intelligent}
Azad~M Madni and Carla~C Madni. 2008.
\newblock Intelligent agents as synthetic role players in scenario-based
  training.
\newblock \emph{Journal of Integrated Design \& Process Science archive},
  12:39--54.

\bibitem[{Mohd et~al.(2023)Mohd, Bravo-Garcia, Love, Gujadhur, and
  Nyadu}]{mohd2023analyzing}
Tauheed~Khan Mohd, Fernando Bravo-Garcia, Landen Love, Mansi Gujadhur, and
  Jason Nyadu. 2023.
\newblock Analyzing strengths and weaknesses of modern game engines.
\newblock \emph{International Journal of Computer Theory and Engineering},
  15(1):54--60.

\bibitem[{Moise(2005)}]{moise_2005_the}
Gabriela Moise. 2005.
\newblock The role of intelligent agents in online learning environment.
\newblock \emph{E-learning and distance learning}.

\bibitem[{Moreno et~al.(2001)Moreno, Mayer, Spires, and
  Lester}]{moreno_2001_the}
Roxana Moreno, Richard~E. Mayer, Hiller~A. Spires, and James~C. Lester. 2001.
\newblock \href {https://doi.org/10.1207/s1532690xci1902_02} {The case for
  social agency in computer-based teaching: Do students learn more deeply when
  they interact with animated pedagogical agents?}
\newblock \emph{Cognition and Instruction}, 19:177--213.

\bibitem[{Park et~al.(2023)Park, O'Brien, Cai, Morris, Liang, and
  Bernstein}]{park_2023_generative}
Joon Park, Joseph O'Brien, Carrie Cai, Meredith Morris, Percy Liang, and
  Michael Bernstein. 2023.
\newblock \href {https://doi.org/10.1145/3586183.3606763} {Generative agents:
  Interactive simulacra of human behavior}.
\newblock \emph{In Proceedings of the 36th annual acm symposium on user
  interface software and technology}, 23.

\bibitem[{Pedersen and Duin(2022)}]{pedersen_2022_ai}
Isabel Pedersen and Ann~Hill Duin. 2022.
\newblock \href {https://doi.org/10.24251/hicss.2022.002} {Ai agents, humans
  and untangling the marketing of artificial intelligence in learning
  environments}.
\newblock \emph{Proceedings of the ... Annual Hawaii International Conference
  on System Sciences}.

\bibitem[{Samuel et~al.(2024)Samuel, Zou, Zhou, Chaudhari, Kalyan, Rajpurohit,
  Deshpande, Narasimhan, and Murahari}]{samuel2024personagym}
Vinay Samuel, Henry~Peng Zou, Yue Zhou, Shreyas Chaudhari, Ashwin Kalyan,
  Tanmay Rajpurohit, Ameet Deshpande, Karthik Narasimhan, and Vishvak Murahari.
  2024.
\newblock Personagym: Evaluating persona agents and llms.
\newblock \emph{arXiv preprint arXiv:2407.18416}.

\bibitem[{Schroeder et~al.(2013)Schroeder, Adesope, and
  Gilbert}]{schroeder_2013_how}
Noah~L. Schroeder, Olusola~O. Adesope, and Rachel~Barouch Gilbert. 2013.
\newblock \href {https://doi.org/10.2190/ec.49.1.a} {How effective are
  pedagogical agents for learning? a meta-analytic review}.
\newblock \emph{Journal of Educational Computing Research}, 49:1--39.

\bibitem[{Skinner(1958)}]{skinner_1958_teaching}
B.~F. Skinner. 1958.
\newblock \href {https://doi.org/10.1126/science.128.3330.969} {Teaching
  machines: From the experimental study of learning come devices which arrange
  optimal conditions for self-instruction.}
\newblock \emph{Science}, 128:969--977.

\bibitem[{Sreedhar et~al.(2025)Sreedhar, Cai, Ma, Nickerson, and
  Chilton}]{sreedhar2025simulating}
Karthik Sreedhar, Alice Cai, Jenny Ma, Jeffrey~V Nickerson, and Lydia~B
  Chilton. 2025.
\newblock Simulating cooperative prosocial behavior with multi-agent llms:
  Evidence and mechanisms for ai agents to inform policy decisions.
\newblock In \emph{Proceedings of the 30th International Conference on
  Intelligent User Interfaces}, pages 1272--1286.

\bibitem[{{\v{S}}turm et~al.(2024){\v{S}}turm, Zajec, {\v{S}}krjanc,
  Mladeni{\'c}, and Grobelnik}]{vsturm2024enhancing}
Jan {\v{S}}turm, Patrik Zajec, Maja {\v{S}}krjanc, Dunja Mladeni{\'c}, and
  Marko Grobelnik. 2024.
\newblock Enhancing cognitive digital twin interaction using an llm agent.
\newblock In \emph{2024 47th MIPRO ICT and Electronics Convention (MIPRO)},
  pages 103--107. IEEE.

\bibitem[{Swan et~al.(2023)Swan, Kido, Roland, and Renato}]{swan_2023_math}
Melanie Swan, Takashi Kido, Eric Roland, and Santos Renato. 2023.
\newblock \href {https://arxiv.org/abs/2307.02502} {Math agents: Computational
  infrastructure, mathematical embedding, and genomics}.

\bibitem[{Wang et~al.(2024{\natexlab{a}})Wang, Chen, Yu, Liu, and
  Jing}]{wang_2024_education}
Chengliang Wang, Xiaojiao Chen, Teng Yu, Yidan Liu, and Yuhui Jing.
  2024{\natexlab{a}}.
\newblock \href {https://doi.org/10.1057/s41599-024-02717-y} {Education reform
  and change driven by digital technology: a bibliometric study from a global
  perspective}.
\newblock \emph{Humanities and Social Sciences Communications}, 11:1–17.

\bibitem[{Wang et~al.(2023)Wang, Dai, Zhu, Teng, and
  Gu}]{wang_2023_understanding}
Chengliang Wang, Jian Dai, Keke Zhu, Yun Teng, and Xiaoqing Gu. 2023.
\newblock \href {https://doi.org/10.1080/10447318.2023.2291609} {Understanding
  the continuance intention of college students toward new e-learning spaces
  based on an integrated model of the tam and ttf}.
\newblock \emph{International Journal of Human-Computer Interaction}, pages
  1--14.

\bibitem[{Wang et~al.(2024{\natexlab{b}})Wang, Wang, Li, Dai, Gu, and
  Yu}]{wang_2024_factors}
Chengliang Wang, Haoming Wang, Yuanyuan Li, Jian Dai, Xiaoqing Gu, and Teng Yu.
  2024{\natexlab{b}}.
\newblock \href {https://doi.org/10.1080/10447318.2024.2383033} {Factors
  influencing university students’ behavioral intention to use generative
  artificial intelligence: Integrating the theory of planned behavior and ai
  literacy}.
\newblock \emph{International Journal of Human-Computer Interaction}, pages
  1--23.

\bibitem[{Wei et~al.(2024)Wei, Wang, Lu, Xu, Liu, Zhao, Chen, and
  Wang}]{wei2024editable}
Yuxi Wei, Zi~Wang, Yifan Lu, Chenxin Xu, Changxing Liu, Hao Zhao, Siheng Chen,
  and Yanfeng Wang. 2024.
\newblock Editable scene simulation for autonomous driving via collaborative
  llm-agents.
\newblock In \emph{Proceedings of the IEEE/CVF Conference on Computer Vision
  and Pattern Recognition}, pages 15077--15087.

\bibitem[{Xia et~al.(2024)Xia, Dittler, Jazdi, Chen, and Weyrich}]{xia2024llm}
Yuchen Xia, Daniel Dittler, Nasser Jazdi, Haonan Chen, and Michael Weyrich.
  2024.
\newblock Llm experiments with simulation: Large language model multi-agent
  system for simulation model parametrization in digital twins.
\newblock In \emph{2024 IEEE 29th International Conference on Emerging
  Technologies and Factory Automation (ETFA)}, pages 1--4. IEEE.

\bibitem[{Yue et~al.(2024)Yue, Lyu, Mifdal, Suh, Zhang, and
  Yao}]{yue2024mathvc}
Murong Yue, Wenhan Lyu, Wijdane Mifdal, Jennifer Suh, Yixuan Zhang, and Ziyu
  Yao. 2024.
\newblock Mathvc: An llm-simulated multi-character virtual classroom for
  mathematics education.
\newblock \emph{arXiv preprint arXiv:2404.06711}.

\bibitem[{Yurtsever et~al.(2020)Yurtsever, Lambert, Carballo, and
  Takeda}]{yurtsever2020survey}
Ekim Yurtsever, Jacob Lambert, Alexander Carballo, and Kazuya Takeda. 2020.
\newblock A survey of autonomous driving: Common practices and emerging
  technologies.
\newblock \emph{IEEE access}, 8:58443--58469.

\bibitem[{Zhang et~al.(2024)Zhang, Zhang-Li, Yu, Gong, Zhou, Liu, Hou, and
  Li}]{zhang_2024_simulating}
Zheyuan Zhang, Daniel Zhang-Li, Jifan Yu, Linlu Gong, Jinchang Zhou, Zhiyuan
  Liu, Lei Hou, and Juanzi Li. 2024.
\newblock \href {https://doi.org/10.48550/arxiv.2406.19226} {Simulating
  classroom education with llm-empowered agents}.
\newblock \emph{arXiv (Cornell University)}.

\bibitem[{Zhao et~al.(2024)Zhao, Huang, Xu, Lin, Liu, and
  Huang}]{zhao2024expel}
Andrew Zhao, Daniel Huang, Quentin Xu, Matthieu Lin, Yong-Jin Liu, and Gao
  Huang. 2024.
\newblock Expel: Llm agents are experiential learners.
\newblock In \emph{Proceedings of the AAAI Conference on Artificial
  Intelligence}, volume~38, pages 19632--19642.

\bibitem[{Zheng et~al.(2025)Zheng, Jiang, Gu, Li, Wang, and
  Zhang}]{zheng2025teaching}
Longwei Zheng, Fei Jiang, Xiaoqing Gu, Yuanyuan Li, Gong Wang, and Haomin
  Zhang. 2025.
\newblock Teaching via llm-enhanced simulations: Authenticity and barriers to
  suspension of disbelief.
\newblock \emph{The Internet and Higher Education}, 65:100990.

\bibitem[{Zhu et~al.(2024)Zhu, Yuan, Wang, Liu, Liu, Deng, Chen, Liu, Dou, and
  Wen}]{zhu_2024_large}
Yutao Zhu, Huaying Yuan, Shuting Wang, Jiongnan Liu, Wenhan Liu, Chenlong Deng,
  Haonan Chen, Zheng Liu, Zhicheng Dou, and Ji-Rong Wen. 2024.
\newblock \href {https://arxiv.org/abs/2308.07107} {Large language models for
  information retrieval: A survey}.

\bibitem[{Zhuge et~al.(2024)Zhuge, Zhao, Ashley, Wang, Khizbullin, Xiong, Liu,
  Chang, Krishnamoorthi, Tian et~al.}]{zhuge2024agent}
Mingchen Zhuge, Changsheng Zhao, Dylan Ashley, Wenyi Wang, Dmitrii Khizbullin,
  Yunyang Xiong, Zechun Liu, Ernie Chang, Raghuraman Krishnamoorthi, Yuandong
  Tian, et~al. 2024.
\newblock Agent-as-a-judge: Evaluate agents with agents.
\newblock \emph{arXiv preprint arXiv:2410.10934}.

\end{thebibliography}

\end{document}